\documentclass[sigconf]{acmart}

\usepackage{hyperref}
\usepackage{caption}
\usepackage{subcaption}
\usepackage{rotating} 
\usepackage{multirow}
\usepackage{float} 
\usepackage{colortbl}                 

\newcommand{\best}[1]{{\cellcolor{green!25}#1}}

\AtBeginDocument{%
  }

\setcopyright{acmlicensed}
\copyrightyear{2026}
\acmYear{2026}
\setcopyright{cc}
\setcctype{by}
\acmConference[MRAC '26]{4th International Workshop on Multimodal, Generative and Responsible Affective Computing}{November 10--14, 2026}{Rio de Janeiro, Brazil}
\acmBooktitle{4th International Workshop on Multimodal, Generative and Responsible Affective Computing (MRAC '26), November 10--14, 2026, Rio de Janeiro, Brazil}
\acmDOI{10.1145/3840474.3840553}
\acmISBN{979-8-4007-2928-7/2026/11}

\begin{document}

\title{When Better Teachers Don’t Make Better Students: Revisiting Knowledge Distillation for CLIP Models in VQA}

\author{Pume Tuchinda}
\affiliation{\institution{VISTEC}\country{Thailand}}

\author{Parinthapat Pengpun}
\affiliation{\institution{Carnegie Mellon University}\country{United State}}

\author{Romrawin Chumpu}
\affiliation{\institution{VISTEC}\country{Thailand}}

\author{Patomporn Payoungkhamdee}
\affiliation{\institution{VISTEC}\country{Thailand}}

\author{Sarana Nutanong}
\affiliation{\institution{VISTEC}\country{Thailand}}

\author{Peerat Limkonchotiwat}
\affiliation{\institution{AI Singapore}\country{Singapore}}
\correspondingauthor
\email{peerat@aisingapore.org}

\renewcommand{\shortauthors}{Tuchinda et al.}

\begin{abstract}
Vision-language models (VLMs) have achieved remarkable success across multimodal tasks, yet their substantial computational demands hinder efficient deployment. 
Knowledge distillation (KD) has emerged as a powerful approach for building lightweight but competitive models, with strong evidence from both language and vision domains. 
However, its application to VLMs, particularly CLIP-style models, remains limited, often constrained to small-scale teachers and narrow evaluation tasks such as classification or retrieval. 
In this work, we present the first systematic study of distillation across a range of CLIP-style teacher models, ranging from standard baselines to large-scale state-of-the-art models. 
Contrary to trends observed in NLP and vision, we find that stronger teachers do not consistently yield better students; in fact, existing distillation frameworks often fail to scale, leading to degraded performance in downstream multimodal tasks such as visual question answering. 
Our findings challenge prevailing assumptions in KD and point toward new directions for designing parameter-efficient multimodal models.
\end{abstract}




\setlength{\textfloatsep}{8pt plus 2pt minus 4pt}  
\setlength{\intextsep}{8pt plus 2pt minus 2pt}     
\setlength{\floatsep}{6pt plus 2pt minus 2pt}      
\maketitle

\section{Introduction}
\label{sec:intro}

Vision language models (VLMs) have demonstrated remarkable capabilities across diverse multimodal tasks, from image captioning to visual question answering (VQA)~\cite{internvl3, cambrian, qwenvl}.
However, the substantial computational demands of these large-scale models limit their practicality for deployment and accessibility.
Creating efficient VLMs that maintain competitive performance while dramatically reducing computational costs remains a critical challenge~\cite{shinde2025surveyefficientvisionlanguagemodels}.

Recent advances in language, reasoning, and vision demonstrate that knowledge distillation (KD) is among the most effective approaches for producing models that are both efficient and high-performing.
For instance, DistilBERT retains 97\% of BERT Base’s performance while reducing the model size by 40\%~\cite{sanh2019distilbert}, and MiniLM showed that students distilled from stronger teachers consistently outperform those distilled from weaker ones~\cite{minilm}.
This pattern holds across multiple works: as teacher models improve, their distilled students also benefit, often without substantial modifications to the distillation methodology~\cite{hinton2015distillingknowledgeneuralnetwork}.
Similar findings extend to vision models; for example, MobileSAM~\cite{mobilesam} matches the performance of the original Segment Anything Model (SAM) while being 60 times smaller.
More recently, works such as TinyLLaMA~\cite{tinyllama} and DeepSeek-R1~\cite{deepseekr1} suggest that directly distilling from powerful teachers can even surpass training from scratch.

Despite KD’s demonstrated successes across multiple domains, distillation for vision-language systems remains underexplored.
In this work, we refer to CLIP-style models as vision encoders and LLaVa-style models as VLMs. 
Existing studies on efficient vision encoders typically distill CLIP-ViT-Tiny from CLIP-ViT-Base and evaluate performance primarily on classification benchmarks such as ImageNet or retrieval tasks like Flickr30k~\cite{clipkd, tinyclip}.
However, these approaches neither examine the scalability of their methods to stronger teachers nor capture the broader utility of the models in downstream multimodal tasks, such as VQA in VLMs.

In this paper, we investigate how well existing distillation techniques for vision encoders transfer across model scales in multimodal tasks.
Our research is centered around the question: \emph{How effective are the current distillation techniques for CLIP-style models, starting from a standard model size to a large-scale model in multimodal tasks?}
Specifically, we adopt the state-of-the-art CLIP-KD~\cite{clipkd} framework and conduct a systematic study spanning multiple teacher-student configurations.

Surprisingly, we find that current methods fail to leverage stronger teachers: students distilled from larger, more performant encoders often underperform or achieve less gain than expected compared to prior literature in a multimodal setting. 
This outcome directly contradicts findings from general KD research, where stronger teachers often yield better students \cite{tinyllava, beyer2022knowledgedistillationgoodteacher, zhang2024studentsteacherdistillingknowledge}.
To better understand this discrepancy, we analyze the underlying dynamics of student-teacher interactions and identify factors that limit scalability in vision-language distillation.

\begin{figure*}[h!]
  \centering
  \includegraphics[width=1\textwidth]{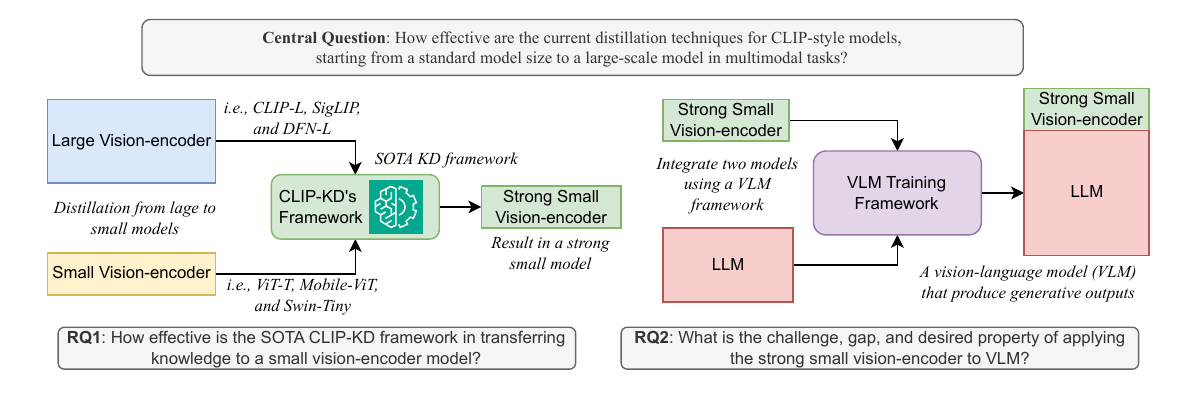}
  \vspace{-8mm}
    \caption{
    Conceptual overview of the framework. The study investigates why existing CLIP-style knowledge distillation methods, despite success in unimodal settings, fail to scale effectively to stronger teachers in multimodal tasks. The study is structured around two questions: (RQ1) knowledge transfer effectiveness in vision encoders, and (RQ2) the challenges of applying distilled encoders within VLMs.
    }
    \vspace{-3mm}
  \label{fig:overview}
\end{figure*}

In summary, our contributions are as follows:
\begin{itemize}
    \item We empirically investigate the effectiveness of existing KD techniques for vision encoders. Contrary to prior studies, stronger teachers fail to improve small students, highlighting a fundamental gap in scalability.
    \item We evaluate parameter-efficient models on multimodal benchmarks and show that, while modest gains appear in certain setups, improvements fail to transfer to tasks such as VQA and Vision Centric tasks, highlighting critical limitations of current KD approaches. 
    \item We study potential bottlenecks in current KD practices, focusing on learning schedules, loss functions, and training data, and provide insights for future improvements.
\end{itemize}

\section{Related Work}
\label{sec:related-work}

\subsection{Large-Scale Vision-Language Models and Scaling Challenges} \label{sec:large_scale}

\noindent\textbf{Evolution of Large-scale Vision-Language Models}
The development of large-scale VLMs has progressed through several architectural paradigms, from early dual-stream models like CLIP~\cite{radford2021learning} to modern autoregressive multimodal language models. Contemporary VLMs such as LLaVA~\cite{liu2023llava}, InstructBLIP~\cite{dai2023instructblip}, and Qwen-VL~\cite{qwenvl} typically employ architectures that combine powerful vision encoders (e.g., CLIP-ViT-Large-336px with 300M+ parameters) with large language models (7B-70B+ parameters), connected through learnable projection layers or more sophisticated cross-modal fusion mechanisms~\cite{zhu2023minigpt4}. 

\noindent\textbf{Scaling Challenges and Computational Bottlenecks}
While these large-scale models demonstrate remarkable capabilities across diverse multimodal tasks, their computational demands present significant deployment challenges. Recent analyses~\cite{shinde2025surveyefficientvisionlanguagemodels} highlight that state-of-the-art VLMs require substantial memory footprints (often exceeding 14GB for 7B models) and exhibit slow inference speeds, limiting their applicability in resource-constrained environments. The scaling behavior of VLMs is particularly complex because performance depends on the careful coordination of both components, unlike unimodal models where scaling laws are better understood~\cite{kaplan2020scaling}.

Furthermore, the training of large VLMs involves multi-stage procedures including vision-language pre-training, instruction tuning, and task-specific fine-tuning~\cite{liu2023improved}, each requiring extensive computational resources. This computational barrier has motivated the research community to explore various efficiency strategies, leading to the emergence of three distinct paradigms for creating practical multimodal systems.

\subsection{Efficiency Paradigms for Vision-Language Models}

Current approaches to improving VLM efficiency can be categorized into three fundamental paradigms, each addressing the scale-performance trade-off from different perspectives ~\cite{Jin2024EfficientML, Shinde2025}.

\noindent\textbf{Data Efficiency Approaches} focus on improving small model performance through enhanced training data curation and optimization strategies. Bunny~\cite{he2024efficientmultimodallearningdatacentric} illustrates this paradigm. While they maintained relatively large vision encoders, they were able to achieve competitive performance using smaller language models (2B-3B parameters) by curating high-quality training data from diverse sources and using careful data selection. Similarly, MM1.5~\cite{Zhang2024MM15MA} and Cambrian~\cite{tong2024cambrian1fullyopenvisioncentric} emphasize the importance of high-quality instruction-following data for multimodal capabilities. These methods typically preserve the core architectural components but optimize the training pipeline through better data quality, curriculum learning~\cite{srinivasan2022curriculumlearningdataefficientvisionlanguage, saha2024exploringcurriculumlearningvisionlanguage}, or data-focused methods~\cite{fang2024vila2vilaaugmentedvila}.

\noindent\textbf{Model Efficiency Approaches} concentrate on architectural innovations and training procedures specifically designed for small-scale models. TinyLLaVA~\cite{zhou2024tinyllavaframeworksmallscalelarge} provides a comprehensive framework for investigating small-scale VLMs, systematically analyzing the effects of different vision encoders, connection modules, and language models in the sub-4B parameter regime. Mipha~\cite{Zhu2024MiphaAC} systematically explored the design choices of smaller VLMs and formulated an architecture that achieved competitive results while keeping data volume consistent. MobileVLM~\cite{chu2024mobilevlmv2fasterstronger} introduces mobile-optimized architectures with lightweight design patterns, while LLaVA-Phi~\cite{zhu2024llavaphiefficientmultimodalassistant} leverages efficient language models like Phi-2 as backbones. These approaches often involve architectural modifications, novel connector designs, and efficient finetuning techniques~\cite{hu2021loralowrankadaptationlarge, han2024parameterefficientfinetuninglargemodels} to maximize performance within parameter constraints.


\noindent\textbf{Knowledge Distillation Approaches} transfer knowledge from large teacher models to compact student models \cite{hinton2015distillingknowledgeneuralnetwork}.
MobileCLIP~\cite{vasu2024mobileclipfastimagetextmodels} introduced a multimodal reinforced training to leverage knowledge transfer from image captioning models, while ACED~\cite{udandarao2025activedatacurationeffectively} showed that active data curation is an effective distillation technique.
CLIP-KD~\cite{yang2024clipkdempiricalstudyclip} demonstrated that distilled vision encoders can preserve the representational strength of larger counterparts. 
However, in all of these works they often limit the scope of their evaluation to classical computer vision benchmarks and do not examine the downstream performance when adapting these multimodal encoders to LLaVA style models.
LLaVa-KD~\cite{cai2025llavakdframeworkdistillingmultimodal} performs joint distillation of both vision encoder and language model, while LLaVA-MoD~\cite{shu2024llavamodmakingllavatiny} leverages mixture-of-expert-based approaches. LLaVADI~\cite{Xu2024LLAVADIWM} systematically explores training strategies, model choices, and distillation algorithms to discover what matters in KD for VLMs.

\section{Proposed Study}
In this paper, we frame the research questions to study and reveal the current gaps and problems in vision-language models using a small vision-encoder model.
%
As illustrated in Figure~\ref{fig:overview}, we formulate two central research questions:
\begin{itemize}
    \item \textbf{RQ1:} How effective is the SOTA CLIP-KD framework in transferring knowledge to a small vision-encoder model?
    \item \textbf{RQ2:} What is the challenge, gap, and desired property of applying the small vision-encoders to VLMs?
\end{itemize}
To answer these questions, we propose three studies as follows.
%
For \textbf{RQ1}, Section~\ref{section:study_1} evaluates the performance of current compact vision encoders compared to larger models across a range of Multimodal benchmarks. 
%
To address \textbf{RQ2}, Section~\ref{subsec:study_2} examines the process of distilling from a large teacher model to a compact student model using an SOTA KD framework.
Furthermore, Section~\ref{subsec:study_3} conducts ablation studies to analyze and probe the knowledge transfer process within the KD framework, offering deeper insights into its limitations and potential improvements.

\section{Effectiveness of Existing Approaches} \label{section:study_1}

\noindent\textbf{Motivation} Recent works such as TinyLLaVA~\cite{tinyllava} and Bunny~\cite{bunny} demonstrate that efficient frameworks can enable compact vision-language models to perform on par with larger models. 
In particular, results from TinyLLaVa and Bunny indicate that using a language model with only 0.5B parameters (Qwen2-0.5B) performs similarly to that of a language model with 8B parameters (Llama-3.1-8B).
However, both approaches still rely on SigLIP-so400m as their default vision encoder, whose parameters are almost similar to those of the language model.
This raises an open question: \emph{to what extent can smaller ViT models serve as viable vision encoders for VQA tasks, and what limitations emerge when compared to their larger counterparts?}

\noindent\textbf{Setup} To explore this, we adopt TinyLLaVA and Bunny as representative frameworks for training efficient vision-language models, with the exact hyperparameters outlined in Table~\ref{tab:tinyllava_hyperparam} and ~\ref{tab:bunny_hyperparam}.
Within each framework, we utilize Qwen2-0.5B and SmolLM-360M as the language models and replace SigLIP-so400m with alternative ViT backbones of varying scales under identical training conditions.
Our study includes CLIP-ViT-L, CLIP-ViT-B, DFN-ViT-L, DFN-ViT-B.
Moreover, we also include a state-of-the-art small vision encoder (CLIP-KD-ViT-T) that was distilled from CLIP-ViT-B on the CC12M dataset using CLIP-KD~\cite{clipkd} to compare the performance of small and large vision models. All vision encoders are trained with the hyperparameters outlined in Table~\ref{tab:clipkd_hyperparam}
For consistency, we evaluate all variants following the Cambrian 1 benchmark~\cite{cambrian}.

\begin{table}[h!]
    \small
    \centering
    \begin{tabular}{lll}
    \toprule
    Hyperparameter & Pretrain & Finetune \\
    \midrule
    Epochs & 1 & 1 \\
    Learning rate & 1e-3 & 2e-5 \\
    Batch size & 16 & 8 \\
    Gradient Accumulation & 4 & 4 \\
    Weight Decay & 0 & 0 \\
    Warmup Ratio & 0.03 & 0.03 \\
    LLM & Frozen & Trained \\
    Vision Encoder & Frozen & Frozen \\
    Adapter & Trained & Trained \\
    Scheduler & Cosine & Cosine \\
    Precision & fp16 & bf16 \\
    Conversation Template & Pretrain & Qwen \\
    Attention & Flash Attention 2 & Flash Attention 2 \\
    Distributed Training & Deepspeed Zero 2 & Deepspeed Zero 3 \\
    \bottomrule
    \end{tabular}
    \caption{TinyLLaVA Hyperparameters}
    \label{tab:tinyllava_hyperparam}
    \raggedbottom
    \vspace{-8mm}
\end{table}

\begin{table}[h!]
    \small
    \centering
    \begin{tabular}{lll}
    \toprule
    Hyperparameter & Pretrain & Finetune \\
    \midrule
    Epochs & 1 & 1 \\
    Learning rate & 5e-4 & 2e-4 \\
    Batch size & 8 & 8 \\
    Gradient Accumulation & 4 & 4 \\
    Weight Decay & 0 & 0 \\
    Warmup Ratio & 0.03 & 0.03 \\
    LLM & Frozen & Trained (LoRA) \\
    Vision Encoder & Frozen & Frozen \\
    Adapter & Trained & Trained \\
    LoRA Rank & -- & 128 \\
    LoRA Alpha & -- & 256 \\
    Projector Learning rate & -- & 2e-5 \\
    Scheduler & Cosine & Cosine \\
    Precision & bf16 & bf16 \\
    Conversation Template & Plain & Plain \\
    Attention & Flash Attention 2 & Flash Attention 2 \\
    Distributed Training & Deepspeed Zero 2 & Deepspeed Zero 3 \\
    \bottomrule
    \end{tabular}
    \caption{Bunny Hyperparameters}
    \label{tab:bunny_hyperparam}
    \raggedbottom
    \vspace{-8mm}
\end{table}

\begin{table}[h!]
    \small
    \centering
    \begin{tabular}{ll}
    \toprule
    Hyperparameter & Value \\
    \midrule
    Epochs & 32 \\
    Learning rate & 5e-4 \\
    Batch size & 256 \\
    Scheduler & Cosine \\
    Precision & fp16 \\
    Warmup Steps & 2000 \\
    CLIP Loss Lambda & 1 \\
    FD Loss Lambda & 2000 \\
    ICL Loss Lambda & 1 \\
    CKD Loss Lambda & 1 \\
    \bottomrule
    \end{tabular}
    \caption{CLIP-KD Hyperparameters}
    \label{tab:clipkd_hyperparam}
    \raggedbottom
    \vspace{-6mm}
\end{table}

\begin{table*}[h!]
\centering
\resizebox{\textwidth}{!}{
\begin{tabular}{l|c|c|cccc|cccc|cccc|cccc}
\toprule
\multirow{2}{*}{\textbf{Method}} 
& &
& \multicolumn{4}{c|}{\textbf{General}} 
& \multicolumn{4}{c|}{\textbf{Knowledge}} 
& \multicolumn{4}{c|}{\textbf{OCR \& Chart}} 
& \multicolumn{4}{c}{\textbf{Vision-Centric}} \\
& \rotatebox{90}{\textbf{ViT Params (M)}}
& \rotatebox{90}{\textbf{Average}} &
\rotatebox{90}{MME$^\text{p}$} &
\rotatebox{90}{MMBench} &
\rotatebox{90}{SEEDBench} &
\rotatebox{90}{GQA} &
\rotatebox{90}{ScienceQA} &
\rotatebox{90}{MMMU} &
\rotatebox{90}{MathVista} &
\rotatebox{90}{AI2D} &
\rotatebox{90}{ChartQA} &
\rotatebox{90}{OCRBench} &
\rotatebox{90}{TextVQA} &
\rotatebox{90}{DocVQA} &
\rotatebox{90}{MMVP} &
\rotatebox{90}{RealWorldQA} &
\rotatebox{90}{CV-Bench$^\text{2D}$} &
\rotatebox{90}{CV-Bench$^\text{3D}$} \\
\midrule
\multicolumn{17}{l}{\textbf{TinyLlaVA}} \\
\multicolumn{17}{l}{\textit{Qwen2-0.5B}} \\
SigLIP-so400m & 428 &\best{38.79} & \best{1,057.64} &\best{66.26} &\best{53.84} &\best{57.08} &53.53 &32.33 &\best{22.80} &40.93 &\best{12.36} &\best{25.40} &\best{45.37} &\best{21.77} &\best{6.67} &\best{31.37} &48.05 &50.08 \\
DFN-ViT-L &303.5 &35.48 &1,018.77 &62.90 &51.45 &53.46 &54.40 &32.33 &21.20 &\best{41.32} &10.08 &7.60 &37.84 &10.47 &5.33 &29.02 &49.10 &50.25 \\
CLIP-ViT-L &303.5 &36.94 &1,046.99 &65.46 &51.85 &55.45 &53.31 &32.78 &22.60 &40.03 &11.48 &12.10 &40.64 &18.33 &3.33 &30.07 &\best{50.35} &\best{51.00} \\
DFN-ViT-B &85.8 &33.77 &912.18 &57.60 &48.27 &51.46 &53.38 &32.33 &22.00 &39.51 &10.32 &4.70 &36.79 &8.01 &5.33 &29.15 &45.83 &50.00 \\
CLIP-ViT-B &85.8 &34.65 &933.43 &61.21 &48.56 &53.39 &\best{54.61} &\best{34.33} &21.90 &40.32 &10.56 &8.10 &36.53 &9.63 &2.67 &29.67 &46.31 &49.92 \\
CLIP-KD-ViT-T &5.5 &31.54 &936.49 &50.77 &42.79 &48.15 &51.83 &32.33 &22.40 &39.86 &9.28 &1.70 &31.19 &6.47 &2.67 &28.63 &39.71 &50.08 \\
\midrule
\multicolumn{17}{l}{\textit{SmolLM-360M}} \\
SigLIP-so400m & 428 &23.12 &550.12 &26.02 &25.29 &42.44 &23.67 &\best{25.67} &13.30 &24.71 &9.56 &2.40 &26.00 &6.84 &4.00 &\best{26.27} &35.05 &51.17 \\
DFN-ViT-L &303.5 &24.30 &514.87 &26.84 &24.24 &46.10 &30.58 &24.67 &\best{19.80} &24.48 &\best{10.36} &4.60 &29.38 &7.01 &\best{4.67} &24.71 &\best{35.33} &50.25 \\
CLIP-ViT-L &303.5 &\best{24.71} &\best{605.14} &26.30 &\best{25.64} &\best{46.28} &22.38 &24.67 &18.30 &24.84 &10.28 &\best{7.80} &\best{32.48} &\best{12.41} &3.33 &25.36 &34.77 &50.25 \\
DFN-ViT-B &85.8 &23.79 &506.93 &\best{27.14} &24.25 &44.43 &\best{35.25} &24.44 &14.00 &23.58 &9.44 &3.00 &27.93 &6.31 &\best{4.67} &25.62 &34.21 &51.08 \\
CLIP-ViT-B & 85.8 &23.77 &517.69 &26.55 &24.36 &44.95 &32.19 &24.56 &15.30 &\best{25.00} &10.12 &4.70 &28.98 &6.38 &0.67 &24.58 &34.70 &\best{51.42} \\
CLIP-KD-ViT-T &5.5 &21.63 &520.53 &25.98 &24.87 &40.56 &23.53 &23.56 &13.20 &23.12 &9.00 &1.00 &21.27 &3.54 &1.33 &24.97 &34.14 &49.92 \\
\midrule
\multicolumn{16}{l}{\textbf{Bunny}} \\
\multicolumn{16}{l}{\textit{Qwen2-0.5B}} \\
SigLIP-so400m &428 &\best{35.60} &1,254.28 &70.85 &\best{55.73} &\best{58.12} &48.83 &29.00 &\best{21.90} &29.57 &\best{10.72} &1.20 &\best{40.39} &8.11 &7.33 &\best{29.67} &45.55 &49.83 \\
DFN-ViT-L &303.5 &34.43 &1,140.73 &67.78 &52.04 &55.33 &52.77 &29.33 &20.30 &31.57 &9.84 &2.20 &36.98 &6.46 &6.67 &26.80 &\best{47.71} &48.08 \\
CLIP-ViT-L &303.5 &35.44 &\best{1,254.50} &\best{70.89} &53.51 &56.68 &50.88 &\best{32.78} &21.30 &34.07 &10.04 &3.40 &38.15 &\best{8.14} &\best{8.67} &27.71 &36.09 &\best{52.00} \\
DFN-ViT-B &85.8 &34.38 &1,115.75 &67.06 &49.70 &53.60 &52.70 &30.67 &21.60 &\best{40.71} &9.32 &2.90 &36.10 &6.27 &8.00 &26.67 &41.72 &47.25 \\
CLIP-ViT-B &85.8 &35.00 &1,153.77 &68.11 &50.54 &54.41 &\best{56.02} &30.22 &21.50 &39.38 &9.72 &\best{4.10} &36.52 &6.62 &7.33 &26.80 &39.43 &51.67 \\
CLIP-KD-ViT-T &5.5 & 31.42 &906.13 &60.49 &43.73 &49.87 &52.98 &30.78 &21.20 &37.89 &9.08 &0.80 &31.19 &5.33 &2.67 &26.14 &38.46 &46.83 \\
\midrule
\multicolumn{16}{l}{\textit{SmolLM-360M}} \\
SigLIP-so400m &428 &\best{27.95} &908.89 &60.83 &\best{32.63} &\best{54.48} &\best{39.31} &23.44 &18.50 &6.12 &8.88 &0.90 &\best{30.45} &6.96 &\best{8.00} &\best{26.01} &35.81 &49.42 \\
DFN-ViT-L&303.5 &27.32 &887.05 &60.45 &27.60 &51.39 &38.08 &27.56 &19.10 &12.47 &8.72 &\best{1.30} &28.86 &5.49 &0.67 &25.10 &35.88 &50.08 \\
CLIP-ViT-L&303.5 &27.37 &\best{973.94} &\best{62.72} &28.43 &51.41 &37.63 &27.00 &19.20 &6.44 &\best{9.00} &1.20 &29.45 &\best{7.13} &0.67 &25.36 &35.05 &48.58 \\
DFN-ViT-B &85.8 &27.06 &932.49 &60.36 &27.01 &49.52 &37.49 &27.00 &17.00 &\best{16.97} &8.40 &0.90 &28.01 &4.96 &2.00 &22.48 &35.12 &49.17 \\
CLIP-ViT-B &85.8 &26.61 &911.71 &61.25 &27.34 &49.97 &36.34 &\best{28.33} &18.80 &6.99 &7.96 &0.60 &27.31 &5.26 &0.67 &23.92 &35.12 &\best{50.33} \\
CLIP-KD-ViT-T &5.5 & 25.84 &841.42 &57.33 &26.39 &45.02 &37.80 &26.44 &\best{19.70} &11.11 &7.92 &0.30 &23.87 &4.23 &1.33 &23.66 &\best{36.37} &49.83 \\
\bottomrule
\end{tabular}}
\caption{Multimodal benchmark performance for pretrained vision encoders, we utilize the patch size 14 for variants for ViT-L families and patch size 16 for ViT-B families. For SigLIP, we utilize the shape-optimized patch size 14 with a 384 resolution variant. For CLIP-KD-ViT-T, we utilized the distilled ViT-Tiny released by CLIP-KD~\cite{clipkd} distilled on CLIP-ViT-B on the CC12M dataset.}
\vspace{-8mm}
\label{tab:prelim}
\end{table*}

\noindent\textbf{SOTA KD frameworks achieve a cost-accuracy balance}
As shown in Table \ref{tab:prelim}, performance decreases as the vision encoder size is reduced, as expected.
For example, in TinyLlava with Qwen2-0.5B, we observe that the performance decreased from 38.79 to 33.77 points when changing from SigLIP-so400m to DFN-ViT-B. 
A similar trend appears in Bunny, where the performance decreased from 35.60 to 34.38.
These results are consistent with previous works showing that larger vision encoders generally yield higher performance.
Yet the efficiency trade-off is quite surprising: the parameter count is reduced by nearly 5$\times$ (428M → 85.8M), while performance drops by only about 5\% in TinyLLaVA and around 1\% in Bunny.
This highlights the strength of smaller vision encoders, which substantially reduces the model size with only modest performance loss.
Beyond parameter efficiency, we also observe runtime improvements, where models with the same size as ViT-B achieve a reduction in inference time of 25\% on average across our evaluations.
%
Together, these findings indicate that, with \textit{the current SOTA VLM frameworks, it is possible to achieve a practical trade-off between efficiency and performance} in compact vision encoders, making them well-suited for real-world applications where both latency and cost are critical factors.

\noindent\textbf{Shrinking 85M$\rightarrow$5M results in less than 5\% drop}
As shown in Table~\ref{tab:prelim}, when we focus on the smallest model, CLIP-KD-ViT-T with only 5.5M parameters, we found that the performance decreased by only $\sim$3\% in Qwen2-0.5B compared to its teacher, CLIP-ViT-B, in TinyLLaVA.
This represents a $\sim$15$\times$ reduction in parameters (85.8M → 5.5M) with less than a 5\% performance drop.
These findings demonstrate that current approaches are highly effective at enhancing the capabilities of compact vision-encoder models.
However, CLIP-KD-ViT-T was distilled from CLIP-ViT-B, a relatively modest-sized teacher.
Given that earlier results showed larger vision encoders consistently yield higher performance, this raises the question of \textit{whether distilling from a larger teacher would yield even better student performance.}
%
Specifically, we could explore distilling from a much larger vision encoder (e.g., 428M parameters) down to an even more compact student model (5.5M parameters), achieving a 99\% parameter reduction while potentially maintaining or improving performance.
This would result in significant cost savings for real-world applications and edge devices.

\section{Gaps and Desired Properties}
\label{subsec:study_2}

Existing literature in knowledge distillation has shown that using a stronger teacher model often yields a better and more robust student model~\cite{hinton2015distillingknowledgeneuralnetwork}.
As demonstrated in Table~\ref{tab:prelim}, our results highlight the potential of using large vision encoders as teachers, rather than traditional medium-sized teachers.
Yet current KD methods for vision encoders typically employ CLIP-ViT-B as the teacher model.
%
%
In Section~\ref{section:study_2_1}, we analyze whether stronger teachers consistently improve student models. 
%
In Section~\ref{section:study_2_2}, we probe the underlying factors that explain the failure in teacher scaling.

%

\begin{table*}[h!]
\centering
\resizebox{\textwidth}{!}{
\begin{tabular}{ll|c|c|cccc|cccc|cccc|cccc}
\toprule
\multicolumn{2}{c}{\textbf{Method}} 
& \multicolumn{1}{c|}{}
&
& \multicolumn{4}{c|}{\textbf{General}} 
& \multicolumn{4}{c|}{\textbf{Knowledge}} 
& \multicolumn{4}{c|}{\textbf{OCR \& Chart}} 
& \multicolumn{4}{c}{\textbf{Vision-Centric}} \\
\textbf{Student} 
& \textbf{Teacher} 
& \rotatebox{90}{\textbf{ImageNet}}
& \rotatebox{90}{\textbf{Average}}
& \rotatebox{90}{MME$^\text{p}$}
& \rotatebox{90}{MMBench}
& \rotatebox{90}{SEEDBench}
& \rotatebox{90}{GQA}
& \rotatebox{90}{ScienceQA}
& \rotatebox{90}{MMMU}
& \rotatebox{90}{MathVista}
& \rotatebox{90}{AI2D}
& \rotatebox{90}{ChartQA}
& \rotatebox{90}{OCRBench}
& \rotatebox{90}{TextVQA}
& \rotatebox{90}{DocVQA}
& \rotatebox{90}{MMVP}
& \rotatebox{90}{RealWorldQA}
& \rotatebox{90}{CV-Bench$^\text{2D}$}
& \rotatebox{90}{CV-Bench$^\text{3D}$} \\
\midrule
\multicolumn{19}{l}{\textit{Qwen2-0.5B}} \\
ViT-T & SigLIP-so400m &25.30&30.41 &823.20 &45.62 &39.30 &44.84 &52.35 &32.11 &21.50 &\best{40.54} &9.16 &1.90 &31.15 &6.23 &1.33 &\best{28.89} &40.19 &50.25 \\
ViT-T & CLIP-ViT-L &34.55&30.29 &784.45 &46.49 &39.35 &44.98 &52.53 &30.56 &22.20 &40.51 &9.24 &1.60 &31.35 &6.17 &2.00 &28.10 &39.71 &\best{50.67} \\
ViT-T & CLIP-ViT-B &36.56 &30.95 &849.02 &48.66 &40.49 &46.41 &\best{52.98} &\best{31.89} &\best{22.80} &39.67 &9.16 &1.90 &31.13 &6.31 &\best{3.33} &26.93 &40.82 &50.33 \\
ViT-T & DFN-ViT-L &37.55&\best{31.34} &\best{910.13} &\best{49.97} &\best{41.77} &\best{47.21} &52.96 &31.78 &22.30 &40.03 &\best{9.28} &\best{2.00} &31.43 &6.12 &2.00 &27.84 &\best{41.03} &50.25 \\
ViT-T & DFN-ViT-B &\best{39.65}&30.45 &813.91 &48.82 &41.52 &46.48 &51.07 &32.11 &21.60 &39.90 &\best{9.28} &1.60 &\best{31.72} &\best{6.43} &1.33 &25.62 &38.73 &50.25 \\
\midrule
Mobile-ViT & DFN-ViT-L &40.70&\best{29.25} &742.64 &43.02 &37.72 &42.15 &\best{51.64} &31.11 &21.00 &\best{38.76} &9.32 &1.50 &\best{30.95} &\best{6.45} &\best{2.67} &\best{26.80} &37.48 &\best{50.25} \\
Mobile-ViT & DFN-ViT-B &\best{41.80} &29.11 &\best{788.61} &\best{43.18} &\best{37.89} &\best{42.62} &49.78 &\best{32.44} &\best{21.80} &36.17 &\best{9.80} &\best{1.70} &29.31 &3.48 &2.00 &26.54 &\best{39.36} &\best{50.25} \\
\midrule
Swin-Tiny & DFN-ViT-L &52.67 &31.43 &855.73 &\best{51.29} &\best{43.31} &\best{49.41} &52.49 &\best{33.00} &\best{21.50} &40.84 &9.32 &1.90 &32.48 &6.66 &\best{2.00} &27.84 &37.83 &\best{50.25} \\
Swin-Tiny & DFN-ViT-B &\best{52.89}&\best{31.65} &\best{862.60} &50.93 &43.00 &49.29 &\best{52.58} &32.78 &\best{21.50} &\best{41.19} &\best{9.80} &\best{2.00} &\best{33.23} &\best{6.82} &1.33 &\best{29.41} &\best{39.22} &50.17 \\
\midrule
ConvNext-Tiny & DFN-ViT-L &52.15&31.51 &730.70 &\best{57.30} &\best{44.91} &\best{48.81} &51.62 &32.56 &21.40 &40.03 &\best{9.68} &1.90 &\best{32.92} &\best{6.70} &2.67 &26.67 &\best{40.19} &\best{50.25} \\
ConvNext-Tiny & DFN-ViT-B &\best{54.04}&\best{31.59} &\best{767.42} &55.65 &43.49 &48.28 &\best{52.53} &\best{32.78} &\best{21.80} &\best{40.64} &9.56 &\best{2.00} &32.48 &6.60 &\best{4.67} &\best{28.63} &37.62 &\best{50.25} \\ 
\midrule
\multicolumn{19}{l}{\textit{SmolLM-360M}} \\
ViT-T & SigLIP-so400m &25.30&22.51 &527.56 &25.95 &24.74 &39.17 &24.50 &24.89 &13.70 &\best{24.61} &9.20 &\best{1.90} &21.27 &\best{5.42} &\best{8.67} &24.18 &\best{35.26} &\best{50.25} \\
ViT-T & CLIP-ViT-L &34.55&\best{22.53} &566.21 &25.98 &25.02 &38.62 &\best{32.49} &23.67 &11.40 &23.90 &9.08 &0.90 &\best{23.47} &4.90 &2.67 &24.71 &35.12 &\best{50.25} \\
ViT-T & CLIP-ViT-B &36.56&\best{22.53} &543.17 &25.18 &24.28 &39.42 &31.60 &23.89 &12.40 &24.55 &9.08 &\best{1.90} &22.32 &4.82 &5.33 &24.05 &34.35 &50.17 \\
ViT-T & DFN-ViT-L &37.55&22.30 &\best{571.33} &26.30 &\best{25.16} &39.58 &25.98 &23.22 &\best{15.50} &24.03 &9.28 &1.80 &22.01 &5.29 &2.67 &24.18 &34.28 &49.00 \\
ViT-T & DFN-ViT-B &\best{39.65} &22.33 &531.20 &\best{26.50} &24.64 &\best{39.91} &25.39 &\best{25.33} &10.80 &24.09 &\best{9.40} &1.80 &23.16 &4.61 &6.00 &\best{24.97} &34.56 &49.50 \\
\midrule
Mobile-ViT & DFN-ViT-L &40.70&\best{22.96} &543.46 &26.00 &\best{24.94} &38.15 &\best{32.09} &\best{26.22} &\best{12.50} &\best{25.36} &\best{9.24} &1.20 &\best{21.74} &4.02 &8.00 &\best{24.97} &\best{34.49} &\best{51.25} \\
Mobile-ViT & DFN-ViT-B&\best{41.80} &22.26 &\best{612.62} &\best{26.59} &24.87 &\best{38.43} &26.64 &26.11 &10.70 &22.57 &8.24 &\best{1.70} &19.69 &\best{4.47} &\best{8.67} &23.14 &34.21 &49.50 \\
\midrule
Swin-Tiny & DFN-ViT-L &52.67&22.92 &\best{559.59} &\best{26.55} &\best{24.32} &\best{41.14} &27.54 &\best{25.67} &\best{19.40} &23.54 &\best{9.60} &1.10 &23.78 &3.79 &2.00 &\best{24.71} &\best{35.26} &\best{50.33} \\
Swin-Tiny & DFN-ViT-B &\best{52.89}&\best{23.16} &528.30 &26.11 &23.65 &41.10 &\best{36.27} &22.33 &14.80 &\best{24.55} &9.24 &\best{1.60} &\best{23.96} &\best{5.08} &\best{6.00} &24.58 &34.63 &50.25 \\
\midrule
ConvNext-Tiny & DFN-ViT-L &52.15&23.04 &502.18 &\best{26.96} &24.70 &\best{42.81} &34.47 &26.00 &\best{13.70} &23.51 &9.56 &\best{1.50} &\best{24.00} &5.11 &2.67 &\best{23.79} &34.56 &50.17 \\
ConvNext-Tiny & DFN-ViT-B &\best{54.04}&\best{23.18} &\best{518.50} &26.23 &\best{24.91} &42.50 &\best{34.76} &\best{26.78} &12.40 &\best{24.51} &\best{9.64} &1.40 &23.36 &\best{5.46} &\best{4.00} &\best{23.79} &\best{34.91} &\best{50.25} \\
\bottomrule
\end{tabular}}
\caption{Multimodal and ImageNet performance for multiple combinations of student and teacher models. Note that we use TinyLlaVa as the vision-text alignment framework for language and vision models.}
\vspace{-8mm}
\label{tab:kd}
\end{table*}

\subsection{Impact of Teacher Strength on Students}
\label{section:study_2_1}

\noindent\textbf{Motivation}
Previous works demonstrate that larger and more capable teacher models generally yield stronger student models~\cite{busbridge2025distillationscalinglaws, Mirzadeh_Farajtabar_Li_Levine_Matsukawa_Ghasemzadeh_2020}.
This trend holds across multiple domains and tasks, where replacing a weaker teacher with a stronger one consistently improves student performance.
Thus, we investigate whether existing CLIP distillation techniques scale to stronger teachers.

\noindent\textbf{Setup} We adopt the CLIP-KD framework, as it integrates multiple distillation strategies and allows us to flexibly experiment with different loss functions, as detailed in Section~\ref{section:study_3_2}.
We experiment on a large-scale setting, including 5 teacher models (SigLIP-so400m, CLIP-ViT-L, CLIP-ViT-B, DFN-ViT-L, and DFN-ViT-B), 4 student models (ViT-T, Mobile-ViT, Swin-Tiny, and ConvNext-Tiny), and 2 language models (Qwen2-0.5B and SmoLM-360M).
We distilled all student models on Conceptual Captions 12M~\cite{cc12m} with the same hyperparameters as CLIP-KD, Table~\ref{tab:clipkd_hyperparam}.
We then adopt the same setup from the previous study (Section~\ref{section:study_1}) and train all the student models with the TinyLLaVA framework. 
Moreover, to further understand the improvement of student models, we also evaluate them on ImageNet~\cite{imagenet}.

\noindent\textbf{Better teachers do not always produce stronger students}
As shown in Table~\ref{tab:kd}, using a stronger teacher often fails to improve student performance across both ImageNet and Multimodal benchmarks. 
In cases where gains occur, they are often limited to a few instances and show no consistent trend when distilling from larger teachers.
%
In the Qwen2-0.5B setup, ViT-T fails to improve in four out of five cases for multimodal benchmarks, while Swin-Tiny and ConvNext-Tiny show no improvement as we distill on larger teachers on either benchmark.
Similarly, for SmolLM-360M, multimodal benchmarks show no significant gains across all student models.
Overall, although some improvements are observed in specific cases, the benefits of distilling from larger teachers remain marginal.

\begin{figure}[h!]
    \centering
    \includegraphics[width=\columnwidth]{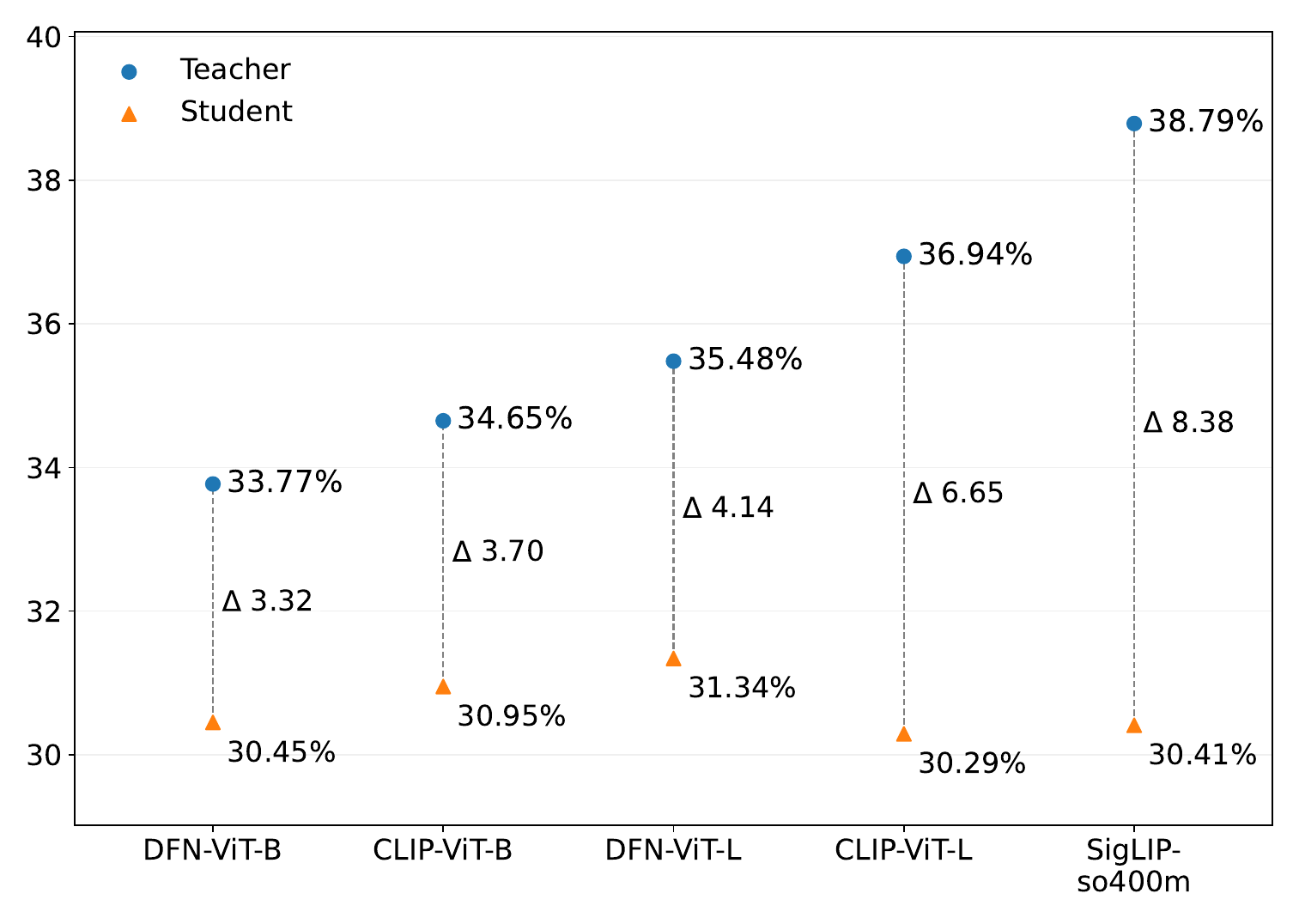}
    \vspace{-9mm}
    \caption{Multimodal performance difference between the various teacher models and the distilled ViT-T as the student model.}    
    \vspace{-8mm}
    \label{fig:kdgap}
\end{figure}

\noindent\textbf{But why do we care about using a larger model as the teacher model?}
As shown in Figure~\ref{fig:kdgap}, the performance difference between the student and teacher widens as the teacher model gets larger.
Distilling from smaller teachers like DFN-ViT-B only yields modest improvements, while larger teachers like CLIP-ViT-L or SigLIP create much bigger performance gaps between teacher and student.
%
For example, the difference in student performance increases from 3.7 when distilled from CLIP-ViT-B to 6.64 when distilled from CLIP-ViT-L, with SigLIP producing the highest overall margin of 8.38.
This larger gap highlights untapped potential: the stronger the teacher, the more headroom exists for the student to improve if the distillation process can effectively transfer the additional capacity.

\vspace{-2mm}

\subsection{Alignment Bottlenecks with Larger Teachers}
\label{section:study_2_2}

\noindent\textbf{Motivation}
As shown in the previous experiment, there were multiple cases where using better/larger teachers did not lead to better results in the student model.
The failure of larger teachers to produce better students contradicts 
the norm in much of the knowledge distillation literature. Therefore, in this experiment, we explore the reason why students failed in the multimodal and ImageNet benchmark to answer the questions from the previous experiment.

\noindent\textbf{Setup}
To understand this, we employ Centered Kernel Alignment (CKA), which measures the similarity between latent representations across different networks~\cite{kornblith2019similarityneuralnetworkrepresentations}. 
In other words, CKA is a way of understanding how well a student model has aligned and captured a teacher model's representations.
CKA has been widely adopted for analyzing representation similarity and is considered a standardized way of measuring distillation fidelity~\cite{csiszárik2021similaritymatchingneuralnetwork,Saha_2022_BMVC,JUNG2023120980}.
Therefore, we compute the CKA between teacher and student representations on the final vision encoder layers across our evaluation tasks. 
Specifically, we focus on teacher-student pairs where larger teachers underperform smaller ones.



\noindent\textbf{Aligning with larger teachers is harder for students} 
Table~\ref{tab:cka} records cases where increasing the teacher model size reduces the downstream performance for the student models. As shown, using the smaller CLIP-ViT-B as a teacher model results in a higher CKA, while the larger teacher model, CLIP-ViT-L, results in a lower CKA. This corresponds to the drop in performance observed across tasks.
The observed decrease in both CKA values and downstream performance when increasing the teacher model size indicates that the student has difficulty aligning with the larger teacher’s representations. This phenomenon can be attributed to a representational capacity mismatch: large vision encoders develop complex and distributed feature spaces optimized for their own scale, which smaller student models are unable to approximate effectively. Consequently, the use of larger teachers may in some cases hinder, rather than improve, student performance.

\begin{table}[h]
\centering
\small
\tabcolsep=0.15cm
\begin{tabular}{l c c r r r}
\toprule
Task &
\begin{tabular}{@{}c@{}}CLIP-ViT-B\end{tabular} &
\begin{tabular}{@{}c@{}}CLIP-ViT-L\end{tabular} &
\multicolumn{1}{c}{\(\Delta\)Perf} &
\multicolumn{1}{c}{\(\Delta\)CKA} \\
\midrule
CV-Bench$^\text{2D}$  & 40.82 / 0.793 & 39.71 / 0.702 & -1.11 & -0.091 \\
GQA        & 46.41 / 0.788 & 44.98 / 0.710 & -1.43 & -0.078 \\
ScienceQA  & 52.98 / 0.929 & 52.53 / 0.902 & -0.45 & -0.027 \\
MMVP       &  3.33 / 0.855 &  2.00 / 0.805 & -1.33 & -0.05 \\
ImageNet   &  30.95 / 0.785 &  30.29 / 0.684 & -0.66 & -0.101 \\
\bottomrule
\end{tabular}
\caption{Tasks where changing the teacher model from \textit{CLIP-ViT-B} $\to$ \textit{CLIP-ViT-L} decreased student (\textit{ViT-T}) performance. Each model cell shows the performance and CKA score in this manner ``Perf (\%) / CKA''. The \(\Delta\)Perf and \(\Delta\)CKA stand for the differences between CLIP-ViT-L and CLIP-ViT-B, and for downstream tasks and CKA score, respectively.}
\vspace{-8mm}
\label{tab:cka}
\end{table}


%




\section{Ablation Studies}
\label{subsec:study_3}
%
As outlined previously, previous works have shown that distilling from stronger models often yields better student models. 
However, from our findings in Section \ref{subsec:study_2}, current distillation techniques for vision encoders struggle to mimic the teacher beyond the ViT-B size.
%
To better understand the limitations, we examine why distilled encoders often fail to scale when applied to VLMs. 
In particular, Section~\ref{section:study_3_1} investigates whether extending the training schedule improves representational alignment.
Section~\ref{section:study_3_2} examines the role of the loss function in shaping student performance.
Section~\ref{section:study_3_3} analyzes whether incorporating additional or better-aligned training data benefits multimodal benchmarks.



\subsection{Effect of Training Duration on Distillation}
\label{section:study_3_1}
\noindent\textbf{Proposed Study}
Previous works~\cite{beyer2022knowledgedistillationgoodteacher} demonstrated that using very long training schedules is a key component for distilling performant student models.
Building on this insight, we investigate whether the same holds for vision-encoder models and examine the extent to which a student can better approximate its teacher given additional training steps.
Following the setup described in Section~\ref{subsec:study_2}, we distill students from DFN-ViT-L and CLIP-ViT-L teachers using training schedules extended from 32 to 96 epochs.
To further assess the effect of longer training, we compute CKA between teacher and student representations, providing a measure of representational alignment.

\begin{figure*}[h!]
    \centering
    
    \begin{subfigure}{0.45\textwidth}
        \centering
        \includegraphics[width=\textwidth]{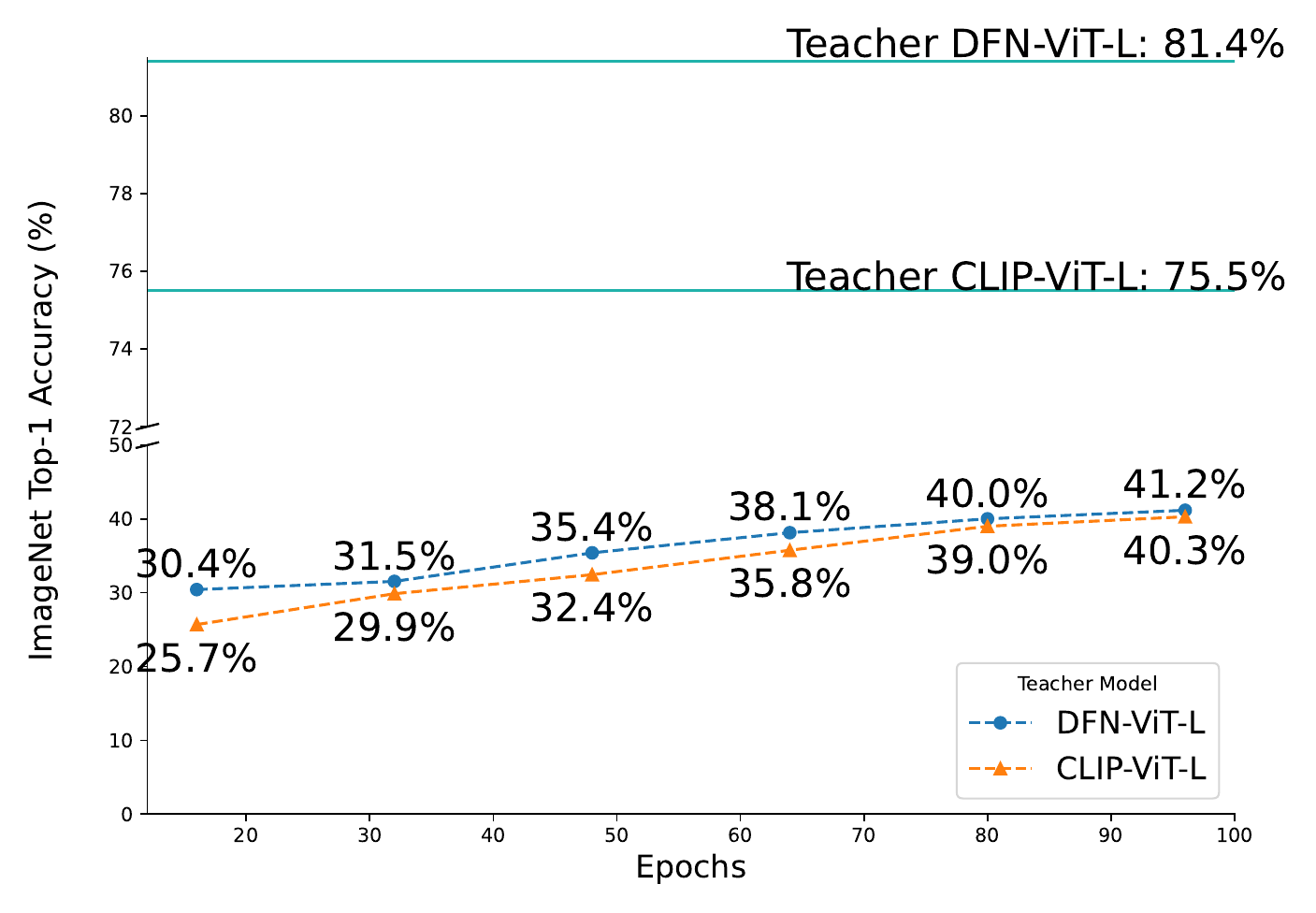}
        \vspace{-7mm}
        \caption{ImageNet performance as the student model (ViT-T) is distilled longer on the CLIP-KD setup.}
        \vspace{-3mm}
        \label{fig:imagenet_longer}
    \end{subfigure}
    \hfill
    \begin{subfigure}{0.45\textwidth}
        \centering
        \includegraphics[width=\textwidth]{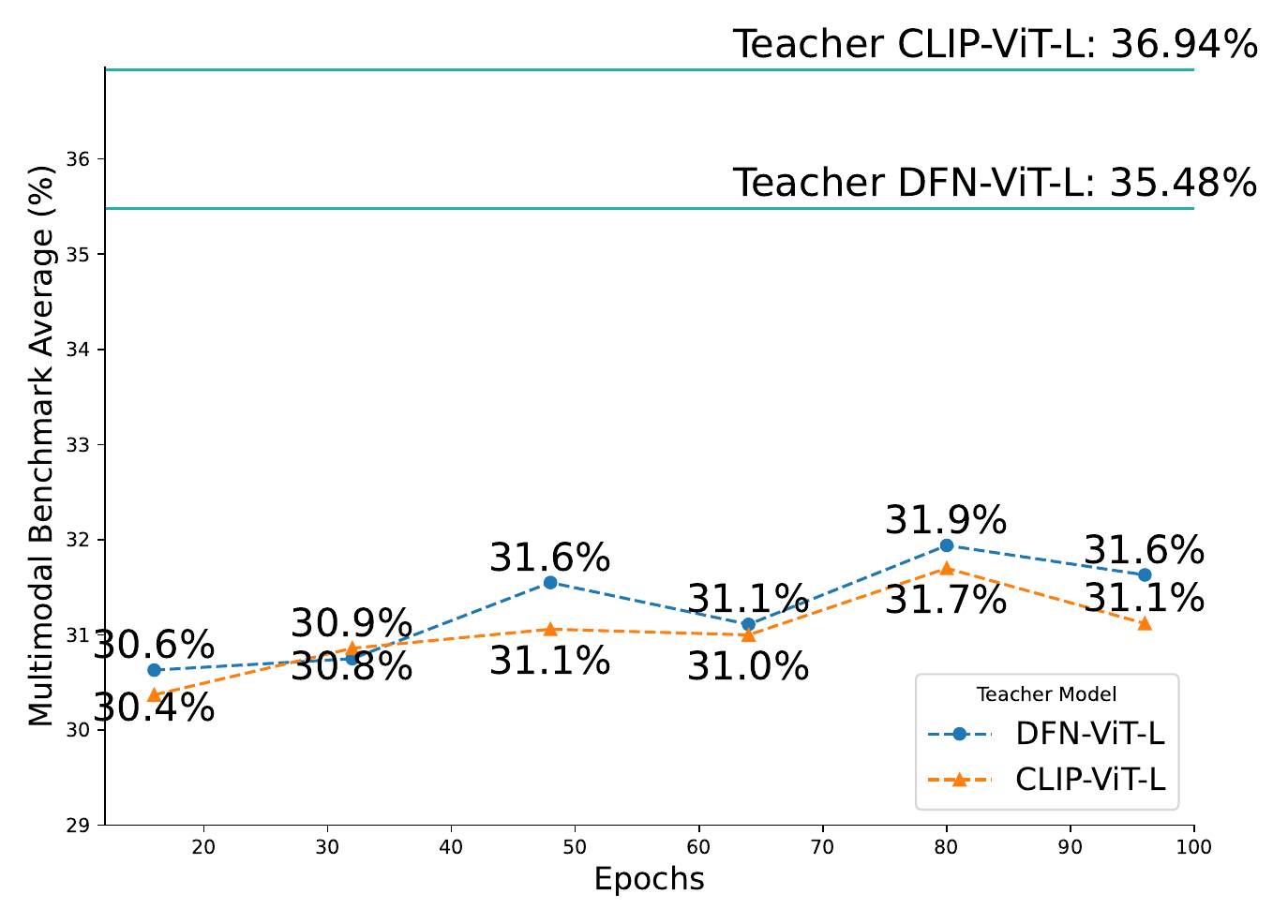}
        \vspace{-7mm}
        \caption{Multimodal performance as the student model (ViT-T) is distilled longer.}
        \vspace{-3mm}
        \label{fig:multimodal_longer}
    \end{subfigure}
    
    \vspace{0.5em} 
    
    \begin{subfigure}{0.45\textwidth}
        \centering
        \includegraphics[width=\textwidth]{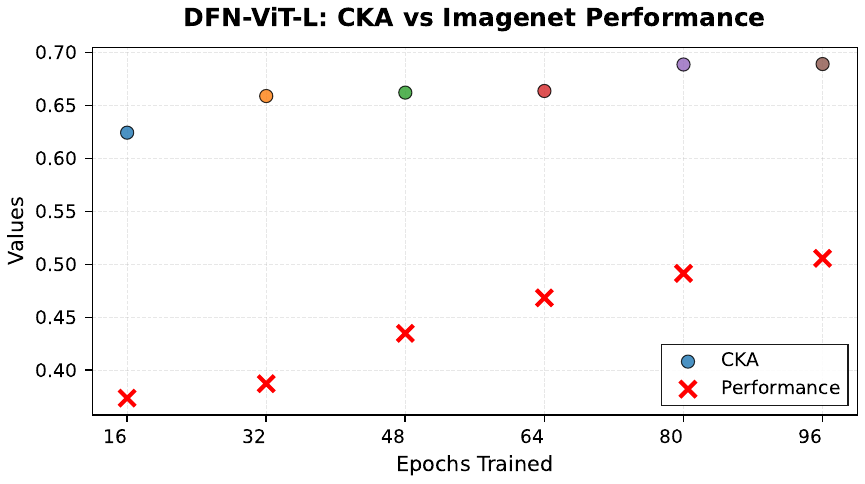}
        \vspace{-7mm}
        \caption{CKA and student performance relative to the teacher for ImageNet across training epochs.  Both metrics increase monotonically as the student model is trained longer.}
        \label{fig:imagenet_cka_distillation}
    \end{subfigure}
    \hfill
    \begin{subfigure}{0.45\textwidth}
        \centering
        \includegraphics[width=\textwidth]{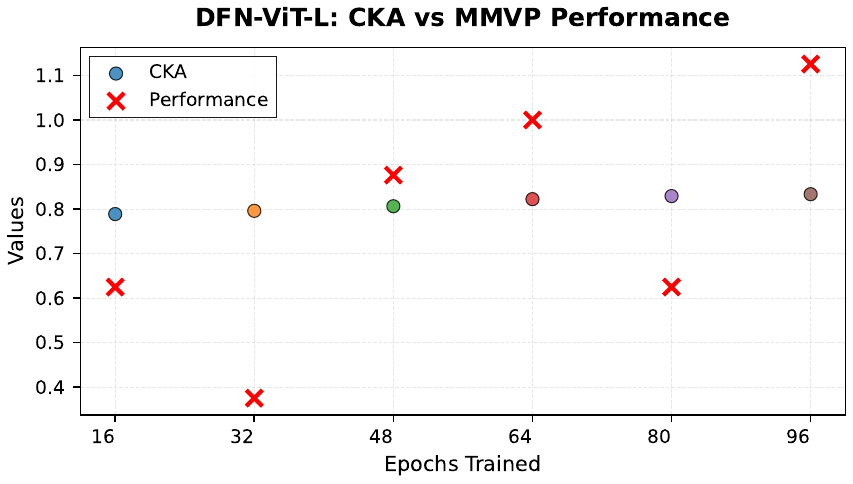}
        \vspace{-7mm}
        \caption{CKA and student performance relative to the teacher for MMVP across training epochs. An erratic pattern can be observed for the performance, while CKA stays relatively constant.}
        \label{fig:mmvp_cka_distillation}
    \end{subfigure}
    \vspace{-3mm}
    \caption{ImageNet and Multimodal performance when distilling the student model (ViT-T) for longer on the CLIP-KD setup. ImageNet accuracy increases steadily with more training epochs (default: 32), whereas Multimodal performance shows only marginal improvements. CKA similarity rises with ImageNet training but remains relatively constant for Multimodal tasks. Results for the Multimodal benchmarks, we take the checkpoints from each epoch and train the vision encoder with the TinyLlaVA framework with Qwen2-0.5B as the LLM.}
    \vspace{-3mm}
    \label{fig:train_longer}
\end{figure*}

\noindent\textbf{Results} As shown in Figure \ref{fig:imagenet_longer} and \ref{fig:multimodal_longer}, extending the distillation schedule from 32 to 96 epochs yields consistent gains on ImageNet classification.
Student models distilled from both DFN-ViT-L and CLIP-ViT-L steadily improve from around 26-30\% top-1 accuracy at 32 epochs to above 40\% by 96 epochs, demonstrating that longer training benefits the student model on image classification tasks.
In contrast, multimodal benchmark performance shows little to no improvement over the same range: average scores fluctuate narrowly between 30-32\%, with no clear upward trajectory.

This is especially pronounced in Figure~\ref{fig:imagenet_cka_distillation} and Figure~\ref{fig:mmvp_cka_distillation}, where we compare the CKA values to the performance of the student model. Since we are interested in the student performance relative to the teacher model, we normalize the performance by dividing the student score by the teacher score.
As shown in Figure~\ref{fig:imagenet_cka_distillation}, both the CKA and the relative performance increase monotonically on ImageNet. On the other hand, Figure~\ref{fig:mmvp_cka_distillation} shows that the CKA on the MMVP task stays consistent, while the relative performance fluctuates wildly between 0.36 and 1.12. This indicates that the distillation process provides no consistent optimization signal for this task.
This divergence highlights that while extended training benefits traditional image tasks, it does not translate into comparable benefits for multimodal tasks.

\subsection{Impact of Loss Function Design}
\label{section:study_3_2}
\noindent\textbf{Proposed Study}
Previous KD studies show that the training objective influences downstream performance~\cite{clipkd, tinyclip}
For example, \citet{tinyclip} reported that switching from a simple MSE loss to contrastive learning~\cite{radford2021learning} boosted ImageNet accuracy of a small model from 19.2 to 53.4.
Motivated by this, we conduct a loss ablation study to assess how different objectives affect multimodal benchmarks.
Specifically, we replace CLIP-KD’s multi-objective loss with MSE~\cite{hinton2015distilling}, Interactive Contrastive Learning~\cite{10.5555/3524938.3525087}, and KL divergence~\cite{fang2021seed,yang2022mutual}.

\noindent\textbf{Results} 
As shown in Table~\ref{tab:loss_ablation}, CLIP-KD’s loss consistently achieves the best performance across benchmarks, in line with prior findings~\cite{clipkd}.
This confirms both the correctness of our setup and the importance of the training objective in KD.
However, even the best-performing loss struggles in multimodal settings, with ViT-B often outperforming ViT-L regardless of the objective.
Taken together, these results highlight a fundamental limitation: existing loss functions fail to scale to more performant teachers on both ImageNet and multimodal benchmarks.
This underscores the need for new loss designs that generalize beyond single-domain benchmarks.

\begin{table}[h!]
\tiny
\centering
\resizebox{\columnwidth}{!}{
\begin{tabular}{ll|c|c}
\toprule
Teacher & Loss & Multimodal & ImageNet \\
\midrule

SigLIP-so400m & CLIP-KD& 30.41 & 25.30 \\
CLIP-ViT-L &CLIP-KD& 30.29 &34.55 \\
CLIP-ViT-B &CLIP-KD& 30.95 &36.56 \\
DFN-ViT-L &CLIP-KD& \best{31.34} &37.55 \\
DFN-ViT-B &CLIP-KD& 30.45 &\best{39.65} \\
\midrule
SigLIP-so400m & MSE & 30.54 &25.06\\
CLIP-ViT-L &MSE &30.20 &34.51 \\
CLIP-ViT-B &MSE &30.28 &\best{35.63} \\
DFN-ViT-L &MSE &30.38 &29.46 \\
DFN-ViT-B &MSE &\best{31.03} &29.70 \\
\midrule
SigLIP-so400M & ICL & 30.33 & 26.92 \\
CLIP-ViT-L &ICL &30.04 &28.59 \\
CLIP-ViT-B &ICL &30.52 &\best{29.04} \\
DFN-ViT-L &ICL &\best{30.66} &26.77 \\
DFN-ViT-B &ICL &30.47 &26.65 \\
\midrule
SigLIP-so400m & KL & \best{30.64} & 30.62 \\
CLIP-ViT-L &KL &30.08 &33.91 \\
CLIP-ViT-B &KL &29.96 &\best{35.04} \\
DFN-ViT-L &KL &30.45 &32.09 \\
DFN-ViT-B &KL &30.56 &27.32 \\

\bottomrule
\end{tabular}}
\caption{Performance comparison of CLIP distillation losses trained on CC12M, showing zero-shot ImageNet classification accuracy and multimodal benchmark results with the TinyLLaVA and Qwen2-0.5B configuration.}
\vspace{-6mm}
\label{tab:loss_ablation}
\raggedbottom
\end{table}

\subsection{Effect of Training Data on Distillation} 
\label{section:study_3_3}

\noindent\textbf{Proposed Study}
Since the data distribution also plays a crucial role in model training~\cite{bunny}, we question whether \emph{do generative datasets help KD's training framework in the multimodal benchmark?}
This is because in many previous KD works, they mainly distill vision encoders on CC-12M, where the distribution of the dataset may not be associated with multimodal tasks.
Moreover, many VLM works demonstrate that the training data is the main factor in VLM~\cite{bunny,cambrian}. 
Therefore, we tested whether adding the pretraining datasets from Bunny and TinyLlaVA would help the student model perform better on multimodal tasks, where the number of samples is increased from 9M to 9M+2M and 9M+0.5M, respectively.
In addition, we also used the same evaluation as in previous experiments. 

\noindent\textbf{Results}
Interestingly, when we find that using DFN-ViT-B as the teacher, adding more data improves the performance from 30.45 to 31.01 points on multimodal benchmarks, as shown in Table~\ref{tab:dataset_ab}.
In contrast, adding Bunny and TinyLlava datasets decreases the performance on the ImageNet. 
We suspect that this is because the distribution of training data is dissimilar to the target benchmark.
On the other hand, we observe negative results for DFN-ViT-L when adding more datasets.
Taken together, these results show marginal improvements, if any, on multimodal benchmarks, suggesting that incorporating pretraining datasets for VLMs may not be the solution.


\begin{table}[h!]
\centering
\resizebox{\columnwidth}{!}{
\begin{tabular}{ll|c|c}
\toprule
Teacher & Dataset & Multimodal & ImageNet \\
\midrule
DFN-ViT-L & CC12M & \best{31.34} &\best{37.55} \\
DFN-ViT-L & CC12M + Bunny Pretrain &31.05 &35.72\\
DFN-ViT-L &CC12M + TinyLlaVA Pretrain &30.38 &33.95 \\
\midrule
DFN-ViT-B & CC12M & 30.45 &\best{39.65} \\
DFN-ViT-B & CC12M + Bunny Pretrain & 30.74 &35.72 \\
DFN-ViT-B &CC12M + TinyLlaVA Pretrain &\best{31.03} &33.52 \\

\bottomrule
\end{tabular}}
\caption{Comparison of DFN-ViT-L and DFN-ViT-B trained with CC12M alone versus CC12M augmented with Bunny or TinyLLaVA pretraining on the CLIP-KD and TinyLlaVA with Qwen2-0.5B setup. While multimodal scores remain similar, ImageNet performance declines with the additional datasets.}
\vspace{-5mm}
\label{tab:dataset_ab}
\end{table}






\section{Discussion, Conclusion, and Future Works}

In this paper, we conduct a series of studies to evaluate current practices for developing small vision encoders and find that current methods do not translate into multimodal performance.
In particular, our paper revisits the CLIP-based knowledge distillation method~\cite{clipkd} and its effectiveness on multimodal settings~\cite{bunny,tinyllava} on a large-scale experiment.

In the first study, Section~\ref{section:study_1}, we show that vision encoders are crucial to VLM performance and that knowledge distillation enables parameter-efficient encoders, reducing parameters from 85.8M to 5.5M with less than a five-point performance drop. 
However, current KD methods typically use Base-size teachers, which may limit small-student performance. This motivates our next question: `Can larger teacher models further improve small models using SOTA KD techniques?'' 
In Section~\ref{section:study_2_1}, we find that while larger teachers offer potential gains (Figure~\ref{fig:kdgap}), existing KD methods fail to scale student performance accordingly. 
Section~\ref{section:study_2_2} shows that this is because students struggle to mimic larger teachers' representations, reflected in lower CKA scores than when distilling from Base-size teachers. This leads us to ask: `Which part of the KD framework is the bottleneck?''

To answer that, we propose three studies to find the bottleneck of each component in the KD method.
In Section~\ref{section:study_3_1}, we found that the current KD method optimizes only for vision-encoder tasks, where we found that there is improvements only on ImageNet, but only a minor improvement in the multimodal task, when we increase the learning time for the KD technique.
Moreover, in Sections~\ref{section:study_3_2} and \ref{section:study_3_3}, we confirm our hypotheses that the current KD method is optimized only for the vision-encoder task, as changing to other settings did not yield any significant improvement.

To summarize, our findings suggest that current KD methods are not well-suited for emerging use cases, such as VLMs, and may require redesign.
Across our experiments, we observe that extending training schedules and employing datasets better aligned with downstream tasks can offer meaningful improvements. 
Moreover, evaluation practices for vision encoders should be updated: instead of focusing solely on traditional benchmarks such as ImageNet or Flickr, modern assessments should also include multimodal tasks.
With the right objectives, training regimes, and data, we believe that small vision encoders hold the potential to approach the performance of their larger counterparts.

Future work should move toward VLM-aware distillation, where compact vision encoders are optimized for downstream multimodal use rather than vision-only teacher imitation. As shown in Section~\ref{section:study_2_1}, stronger teachers do not consistently yield stronger students, and Section~\ref{section:study_2_2} suggests that this may stem from representation mismatch between large teachers and small students. Moreover, the ablations in Section~\ref{subsec:study_3} show that longer training, alternative losses, and additional data are not sufficient to reliably improve multimodal performance. These findings motivate future work on task-aware objectives, capacity-aware transfer, and VLM-based evaluation protocols.

\begin{acks}
This research is supported by the National Research Foundation, Singapore, under its National Large Language Models Funding Initiative. Any opinions, findings, and conclusions or recommendations expressed in this material are those of the author(s) and do not reflect the views of the National Research Foundation, Singapore.
\end{acks}



\bibliographystyle{ACM-Reference-Format}
\bibliography{sample-base}





\end{document}